\documentclass{article}

\usepackage[utf8]{inputenc} 
\usepackage[T1]{fontenc}    
\usepackage{booktabs}       
\usepackage{amsfonts}       
\usepackage{nicefrac}       
\usepackage{microtype}      
\usepackage{graphicx}
\usepackage{enumerate}
\usepackage{caption}
\usepackage{subcaption}
\usepackage{array}

\usepackage{hyperref}

\usepackage{color,soul}
\usepackage[nomargin,inline,marginclue,draft]{fixme}

\usepackage{amsmath,amssymb}

\newcommand{\bm}[1]{#1}
\newcommand{\tens}[1]{\bm{#1}}
\newcommand{\rulesep}{\unskip\ \vrule width 0.75pt\ }

\usepackage[accepted]{icml2019}

\icmltitlerunning{Temporal Gaussian Mixture Layer for Videos}

\begin{document}
\twocolumn[
\icmltitle{Temporal Gaussian Mixture Layer for Videos} 

\icmlsetsymbol{equal}{*}

\begin{icmlauthorlist}
\icmlauthor{AJ Piergiovanni}{iu}
\icmlauthor{Michael S. Ryoo}{iu}
\end{icmlauthorlist}

\icmlaffiliation{iu}{Department of Computer Science, Indiana University}

\icmlcorrespondingauthor{AJ Piergiovanni}{ajpiergi@indiana.edu}
\icmlcorrespondingauthor{Michael Ryoo}{mryoo@indiana.edu}

\icmlkeywords{Activity detection}

\vskip 0.3in
]

\printAffiliationsAndNotice{}  

\begin{abstract}
We introduce a new convolutional layer named the \emph{Temporal Gaussian Mixture} (TGM) layer and present how it can be used to efficiently capture longer-term temporal information in continuous activity videos. The TGM layer is a temporal convolutional layer governed by a much smaller set of parameters (e.g., location/variance of Gaussians) that are fully differentiable. We present our fully convolutional video models with multiple TGM layers for activity detection. The extensive experiments on multiple datasets, including Charades and MultiTHUMOS, confirm the effectiveness of TGM layers, significantly outperforming the state-of-the-arts\footnote{Code/models: \href{https://github.com/piergiaj/tgm-icml19}{https://github.com/piergiaj/tgm-icml19}}.

\end{abstract}

\section{Introduction}

Activity videos are spatio-temporal data: they are image frames with a specific width/height (XY) concatenated along time axis (T). Recognition from such videos requires capturing both spatial and temporal information, desirably using learned convolutional kernels. Temporal convolution is particularly beneficial in activity `detection' tasks, which require making activity decisions at every frame given a \emph{continuous} video \citep{sigurdsson2016hollywood,yeung2015every}. Previous methods investigated using 3-D XYT convolutional filters \citep{tran2014c3d,carreira2017quo} as well as the models with 2-D XY conv. layers followed by 1-D temporal conv. \citep{tran18closer}, pooling or attention layers \citep{piergiovanni2017learning}.

Understanding complex multi-activity videos requires capturing information in long-term time intervals. Different frames contain different information, and the model needs to learn to take advantage of as many frames as possible, while abstracting them efficiently. Previous attempts of simply pooling representations over time or learning temporal conv. filters with a small number of frames (e.g., 16 or 64) was thus often insufficient to fully consider rich long-term temporal context. Simultaneously, bruteforcely increasing the temporal filter length (to look at more frames) results more learnable parameters, requiring more training data, which can be expensive when activities are rare.

\begin{figure}
    \centering
    \includegraphics[width=\linewidth]{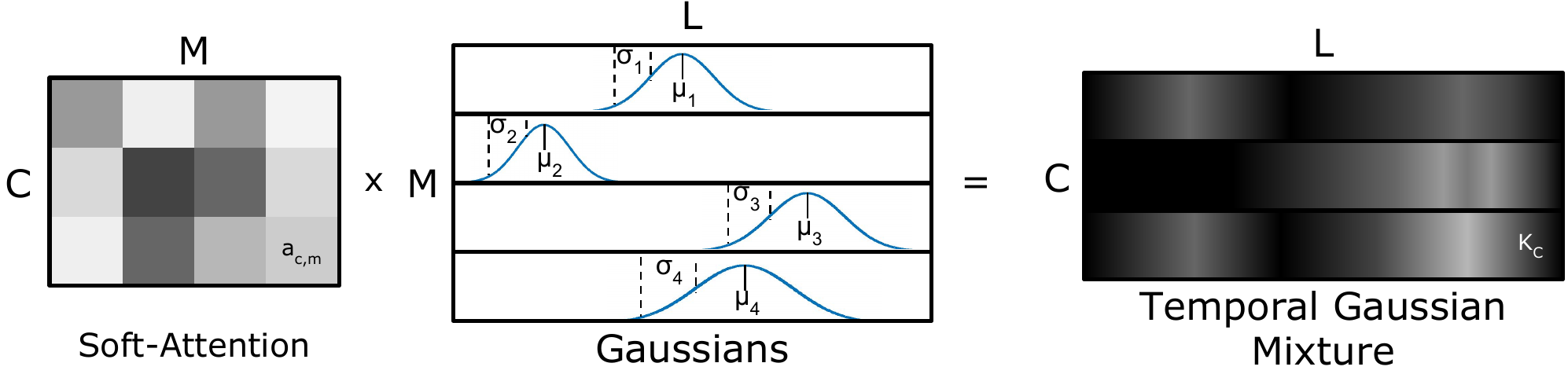}
    \caption{Example illustrating how our Temporal Gaussian Mixture layer is computed. Multiple ($M$) temporal Gaussian distributions are learned, and they are combined with the learned soft attention (mixing) weights to form the $C$ temporal convolution filters. $L$ is the temporal length of the filter.}
    \label{fig:mixture}
\end{figure}

In this paper, we introduce a new convolutional layer named the \emph{Temporal Gaussian Mixture} (TGM) layer, and present how it can be used to efficiently capture longer-term temporal information in activity videos. Our temporal Gaussian mixture layer is a temporal convolutional layer, whose filters/kernels are controlled by a set of (temporal) Gaussian distribution parameters. Each of our temporal Gaussian distributions specify (temporally) `where' the model should look, and our Gaussian mixture layer combines them as multiple convolutional filters to be applied on top of temporally-continuous representations. This layer allows the video representation at each time step to be constructed while focusing on different neighboring temporal regions, instead of only focusing on its local segment. It is a convolutional layer governed by a much smaller set of parameters (i.e., locations/variances of the Gaussians as well as their mixture weights) that are fully differentiable. Importantly, the number of parameters of the TGM layer is independent of the length of the filter, allowing the model to capture longer-term temporal information without additional parameters.

The motivation behind our temporal Gaussian mixture layer is to learn the temporal structure of an activity as a composition of temporal Gaussian regions/attentions. Such structure allows the model to obtain a compact spatio-temporal representation abstracting each (long-term) time interval, using multiple temporal conv. layers with far fewer parameters. 
It is also related to the previous temporal attention works \citep{piergiovanni2017learning}, but our model is designed to be fully convolutional to handle continuous data and it learns more compositional structures with multiple layers.


We present video-CNN models using our TGM layers for activity detection in continuous videos. Our model stacks TGM layers on top of several state-of-the-art CNNs such as I3D \citep{carreira2017quo}. This enables our model to capture longer-term temporal information than provided by the base CNNs, compositionally modeling the temporal structure with multiple TGM layers.
Our model was evaluated on multiple public datasets including MultiTHUMOS and Charades, and was able to outperform the best previous activity detection CNNs by a meaningful margin.

\section{Related Works}

Learning video representations for human activity recognition has been successful. CNN methods allow end-to-end learning of video features and representations optimized for the training data, performing superior to traditional works \citep{aggarwal11} for video understanding.

Two-stream CNN models take a single RGB frame and a small number of optical flow frames as inputs to capture both motion and appearance information in videos~\citep{simonyan2014two,feichtenhofer2016convolutional}. Models learning 3-D spatio-temporal (XYT) convolutional filters were designed and applied to many activity recognition tasks as well~\citep{tran2014c3d,carreira2017quo,tran2017convnet,hara2017learning}. Large scale datasets for activity detection, such as THUMOS~\citep{THUMOS14}, ActivityNet~\citep{caba2015activitynet}, Kinetics~\citep{kay2017kinetics}, and Charades~\citep{sigurdsson2016hollywood} provided these approach the necessary training data to learn the models. Such 3-D XYT CNNs were also used to capture spatio-temporal information for activity detection~\citep{xu2017r,shou2016temporal,shou2017cdc,zhao2017temporal}. However, all these CNNs were limited to the consideration of a fixed local video segment (e.g., 16 frames in \citep{tran2014c3d} and 64-99 frames in \citep{carreira2017quo}) when making activity decisions.



Some works studied combining representations over longer-term temporal intervals ~\citep{karpathy2014large,ng2015beyond,varol17}, but it was generally done with a temporal pooling of local representations or spatio-temporal convolutions with slightly larger fixed intervals. Recurrent neural networks (RNNs) have also been used to model activity transitions between frames~\citep{yeung2015every,yeung2016end,escorcia2016daps}, but they were strictly sequential and had limitations in maintaining temporal information over a longer temporal duration, particularly for videos with multiple complex activities. Recently, CNN models using temporal attention for activity videos \citep{piergiovanni2017learning,piergiovanni2018super} were studied as well. However, a fully convolutional model to analyze continuous videos while efficiently representing information in long term intervals has been lacking.

Several works have explored using parameterized convolutional kernels for point clouds \cite{xu2018spidercnn} or images \cite{cohen2018spherical}, but were quite different from our design to handle long-term temporal information.

Our layer is different from the previous standard spatio-temporal convolutional layers in that it relies on significantly fewer parameters by forcing filter shapes to be Gaussian compositions. Our temporal layer is also different from previous Gaussian Mixture Model layers~\citep{variani2015gaussian} in that our layer is convolutional while they are not.

\section{Approach}

In this section, we introduce a new convolutional layer named the \emph{Temporal Gaussian Mixture} (TGM) layer, and present how it can be used for activity recognition. Our Temporal Gaussian Mixture layer is a temporal convolutional layer to be applied on top of a sequence of representations (usually from frame-level or segment-level CNNs), whose filters/kernels are controlled by a set of (temporal) Gaussian distribution parameters. The motivation is to make each temporal Gaussian distribution specify (temporally) `where to look' with respect to the activity center, and represent the activity as a collection/mixture of such temporal Gaussians convolved with video features. Our layer is fully differentiable and trainable using standard backpropagation.

Our TGM layer can be interpreted as a a form of 1-D convolution where the filters are determined by a mixture of Gaussians. However, we design our TGM layer differs from the standard temporal convolutional layers of learning 1-D (time) or 2-D (channel-by-time) filters in the following aspects (illustrated in Fig. \ref{fig:various-baselines}):

\vspace{-3pt}
\begin{enumerate}

\item Instead of learning temporal convolution filters of any arbitrary values, our filter is forced to have the form of a temporal Gaussian mixture shared across all frame-level channels. This allows the layer to rely on significantly fewer number of (fully differentiable) parameters, while capturing the concept of temporal structure/attention.

\item Our temporal Gaussian mixture layer handles multiple 3-D tensors internally to preserve channels from the frame-level CNN by adding a new temporal channel axis. Its input is 3-D (channel-by-channel-by-time), where one channel dimension is inherited from the frame-level CNN and this dimension size remains unchanged. This is a form of grouped convolution, and we call this more specifically temporal channel grouping (TC-grouping).

\end{enumerate}
\vspace{-3pt}

\subsection{Temporal Gaussian Mixture layer}
\begin{figure}
\centering
  \begin{subfigure}{.23\textwidth}
  \centering
    \includegraphics[width=\textwidth]{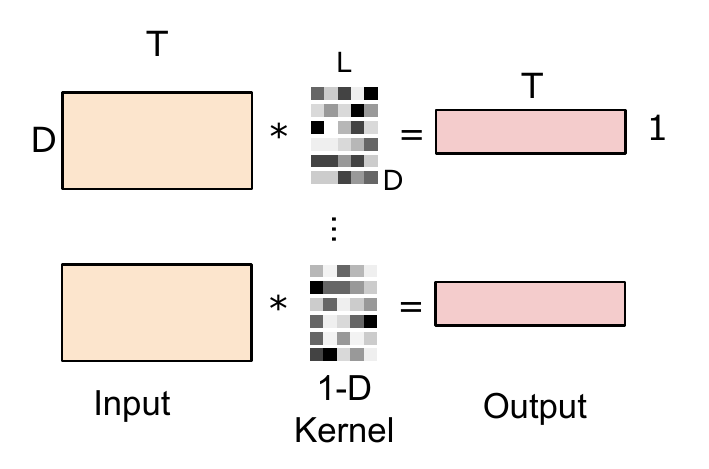}
    \caption{}
    \label{fig:conv1d}
  \end{subfigure}\rulesep%
  \begin{subfigure}{.23\textwidth}
  \centering
    \includegraphics[width=\textwidth]{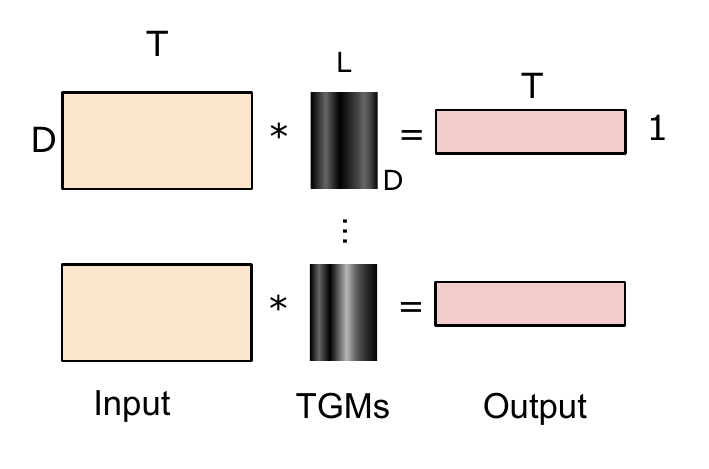}
    \caption{}
    \label{fig:tgm-shared-1d}
  \end{subfigure}
  
  \begin{subfigure}{.23\textwidth}
  \centering
    \includegraphics[width=\textwidth]{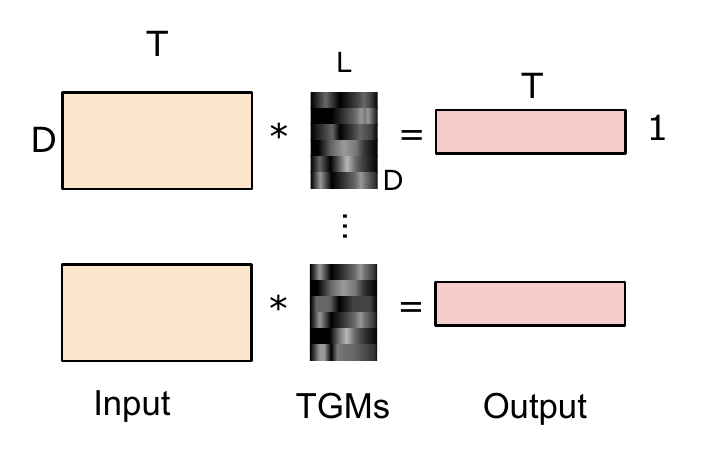}
    \caption{}
    \label{fig:tgm-as-1d}
  \end{subfigure}\rulesep%
  \begin{subfigure}{.23\textwidth}
  \centering
    \includegraphics[width=\textwidth]{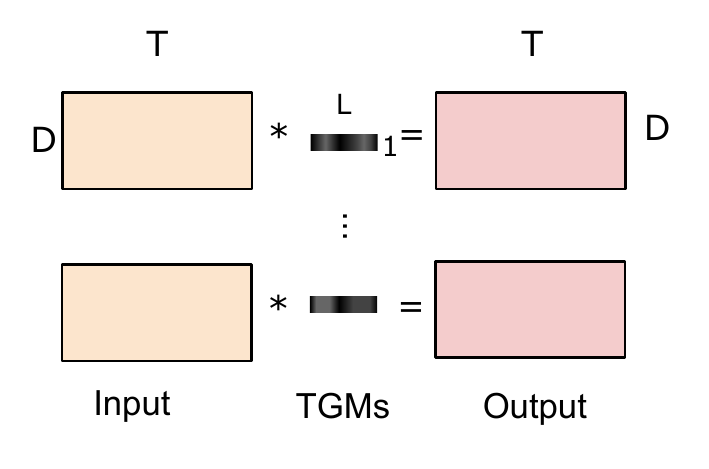}
    \caption{}
    \label{fig:real-tgm}
  \end{subfigure}
  \caption{{\bf (a-c)} Different forms of 1-D temporal convolutions which take a $D\times T$ input and produces a $C\times T$ output based on $C$ number of $D\times L$ kernels: {\bf (a)} the standard 1-D convolution, {\bf (b)} using Gaussian mixtures for 1-D convolution while sharing Gaussian mixtures across input channels, and {\bf (c)} using $D$ different Gaussian mixtures for 1-D convolution. {\bf (d)} Our TGM layer (with TC-grouping) in its simplest form (i.e., 1-layer case) applying the $1\times L$ temporal kernel in a 2-D convolutional fashion, maintaining both time and feature axis.}
  \label{fig:various-baselines}
\end{figure}

Our temporal Gaussian mixture layer takes a 3-D input with the dimensionality of $C_{in}\times D\times T$, where $C_{in}$ is the number of input channels, $D$ is the dimensionality of the representations from frame-level (or segment-level) CNNs, and $T$ is the time. Given such input, the TGM layer convolves it with $C_{out}$ number of $1\times L$ filters/kernels, generating a $C_{out} \times D \times T$-dim representation as an output. $L$ is the temporal length of the temporal Gaussian mixture filter. $D$ is usually 1K or 4K and $T$ is the number of time steps (frames) in each video (i.e., it varies per video). $C_{out}$ is the number of different mixtures, corresponding to the number of output channels in standard convolution. In Fig. \ref{fig:various-baselines}, we compare our TGM layer with the different forms of temporal convolutions.
We experimentally compare all these versions to confirm the benefits of our TGM layer.


Our layer is composed of a set of $M$ Gaussians. Each Gaussian has 2 parameters: a center $\hat{\mu}$ and a width $\hat{\sigma}$. Each layer has additional hyper-parameters: $L$, the temporal duration and $M$, the number of Gaussians to learn.  We constrain the learned center to be between 0 and $L$ and $\sigma$ to be positive:
\begin{equation}
\begin{split}
    \mu = (L-1)\cdot \frac{\tanh{(\hat{\mu}})+1}{2}, ~~\sigma^2 = \exp{(\hat{\sigma})}.
\end{split}
\end{equation}
We use the above $\mu$ and $\sigma$ to construct the temporal Gaussian kernels. This acts as a strong sparsity constraint on the convolutional kernel as well as a drastic reduction of the number of learnable parameters. We construct a temporal Gaussian mixture convolutional kernel as:
\begin{equation}
    \hat{\bm{K}}_{m,l} = \frac{1}{Z}\exp{-\frac{(l-\mu_m)^2}{2\sigma_m^2}}
\end{equation}
where $Z$ is a normalization constant such that $\sum_l^L \hat{K}_{m,l} = 1$, resulting in $\hat{\bm{K}}$ being an $M\times L$ matrix.

Instead of making the model learn a separate set of Gaussian distributions per activity class, we take the approach of maintaining multiple Gaussian distributions shared across classes and obtain a Gaussian `mixture' filter by learning soft-attention weights. We learn a set of soft-attention weights per output channel $i$, $\bm{\omega}\in\mathcal{R}^{C_{out}\times M}$. We create the soft-attention weights by applying the softmax function over the $M$ Gaussians, enforcing each input channel weights sum to 1.
\begin{equation}
    a_{i,m} = \frac{\exp{\bm{\omega}_{i,m}}}{\sum_j \exp{\bm{\omega}_{i,j}}}
\end{equation}

Based on temporal Gaussian distributions $\hat{\bm{K}}_{i}$ and attention weights $a_{i,m}$, the temporal convolution filters our TGM layer is computed as:
\begin{equation}
    \bm{K}_i = \sum_m a_{i,m} \hat{\bm{K}}_i.
\end{equation}
This provides us convolutional filters having the form of a mixture of temporal Gaussians, controlled based on $2\cdot M + C_{in}\cdot C_{out}\cdot M$ parameters (instead of learning $D^2\cdot L$ parameters without any constraint, as in standard temporal convolution where $C << D$). An overview of this process is shown in Fig.~\ref{fig:mixture}.

\subsubsection{Single TGM layer - direct per-class activity modeling}
\begin{figure}
    \centering
    \includegraphics[width=\linewidth]{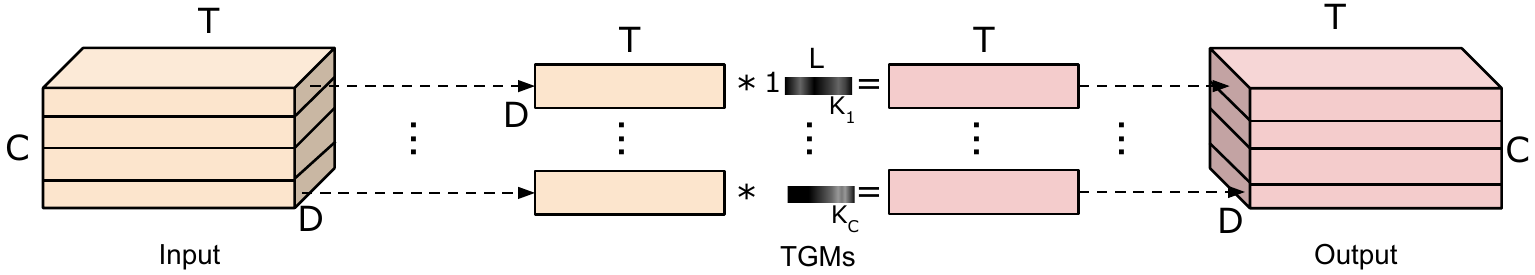}
    \caption{Illustration of a TGM layer with grouped convolution. This layer learns a set of $C$ Gaussian mixtures that are convolved with the input channels.}
    \label{fig:per-class}
\end{figure}

\begin{figure*}
    \centering
    \includegraphics[width=\textwidth]{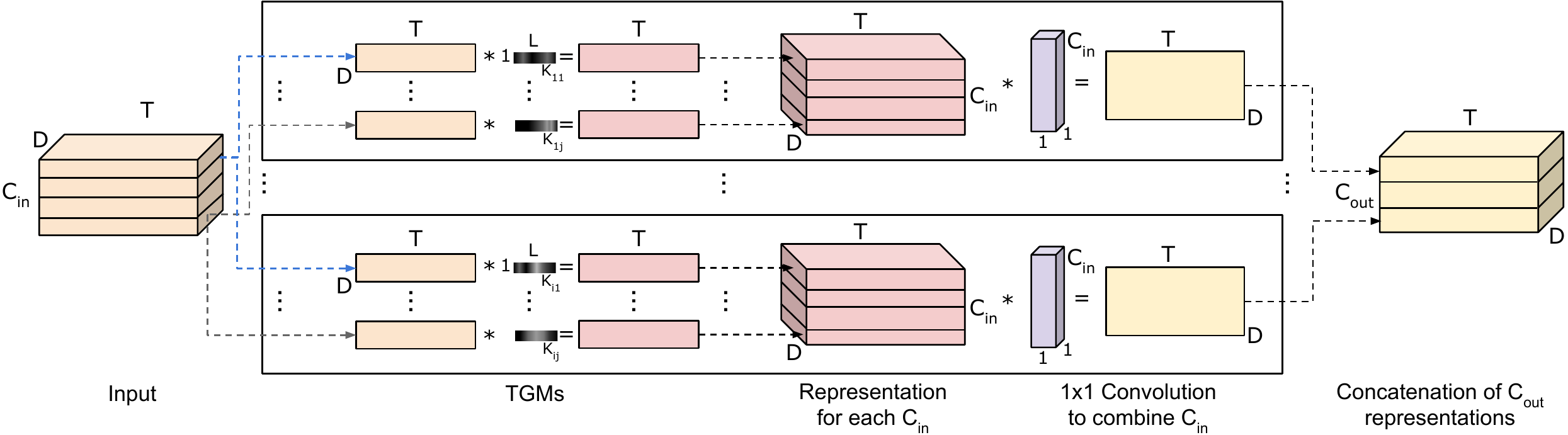}
    \caption{Illustration of a TGM layer with channel combination. The kernels are applied to each input channel, $C_{in}$, and a 1x1 convolution is applied to combine the $C_{in}$ input channels for each output channel, $C_{out}$.}
    \label{fig:shared}
\end{figure*}

The representation we obtain by applying our base CNNs to each frame (or local segment) has the dimensionality of $D$, and stacking them along time axis provides us the representation with $1\times D\times T$-dim. That is, in the case of using only one TGM layer to capture activity representations, our $C_{in}$ is fixed to $1$ and $C_{out}$ is fixed to be the number of activity classes. This is the simplest case of our model, attaching one TGM layer on top of the $1\times D\times T$ representation.

Our convolutional kernel, $\bm{K}$, has a learned Gaussian mixture for each activity class. Let the video features $\bm{v}$ be a $D\times T$ matrix. Each $\bm{K}_i$ is a 2-D convolutional filter with a size of $1\times L$, and convolving this with $\bm{v}$ provides us a representation $\tens{S}$ with $C_{out}$ number of $D \times T$ responses since $C_{in}$ is 1 in this case.
This per-class representation can then be used as input to a fully-connected layer for activity classification. For $i\in \{1,2,\ldots,C_{out}\}$:
\begin{equation}
    s_i = \bm{v} * \bm{K}_i, ~~\tens{S}=[s_1,s_2,\ldots,s_{C_{out}}]
\end{equation}
Fig. \ref{fig:various-baselines} visually illustrates how each TGM filter is convolved with the input (Fig.~\ref{fig:real-tgm}), compared to the standard 1-D convolution (Fig.~\ref{fig:conv1d}) or other forms of the temporal layers (Fig.~\ref{fig:various-baselines}b-c).

\subsubsection{Multiple TGM layers - grouped convolution}

We generalize the above formulation to allow the TGM layers to be sequentially applied. The idea is to enable our model to capture more complex, nonlinear temporal structure by having multiple levels of temporal layers. In this case, the input for each layer is $C_{in}\times D\times T$ dimensional (instead of $1\times D\times T$), where the input channels are the number of output channels from the previous layer. Our kernels at each layer, $\bm{K}_i$, are parameterized and learned as before.

By using grouped convolution with the number of groups set to $C_{in}$, we can efficiently separate the input into per-channel values and convolve each of them with the designated $\bm{K}_i$ kernel, as shown in Fig.~\ref{fig:per-class}. That is, we learn a filter $\bm{K}_i$ per channel by setting $C_{in} = C_{out}$. For $i\in [1,C_{out}]$,

\begin{equation}
    s_i = \tens{f}_i * \bm{K}_i, ~~\tens{S} = [s_1, s_2, \ldots s_{C_{out}}]
\end{equation}
Here, $\tens{f}$ is a $C_{in}\times D\times T$ tensor, where $D$ is the dimensionality of the feature and $T$ is the number of frames. The result of the per-channel convolution, $s_i$, is a $D\times T$ representation. We concatenate these representations along the channel axis, resulting in $\tens{S}$, a $C_{out}\times D\times T$ representation. As this convolution results in the same output shape, we can stack these layers. Each layer is able to capture increasing temporal resolution, allowing the model to capture levels of abstractions.

\subsubsection{Multiple TGM layers - channel combination}
\begin{figure*}
    \centering
    \includegraphics[width=0.99\textwidth]{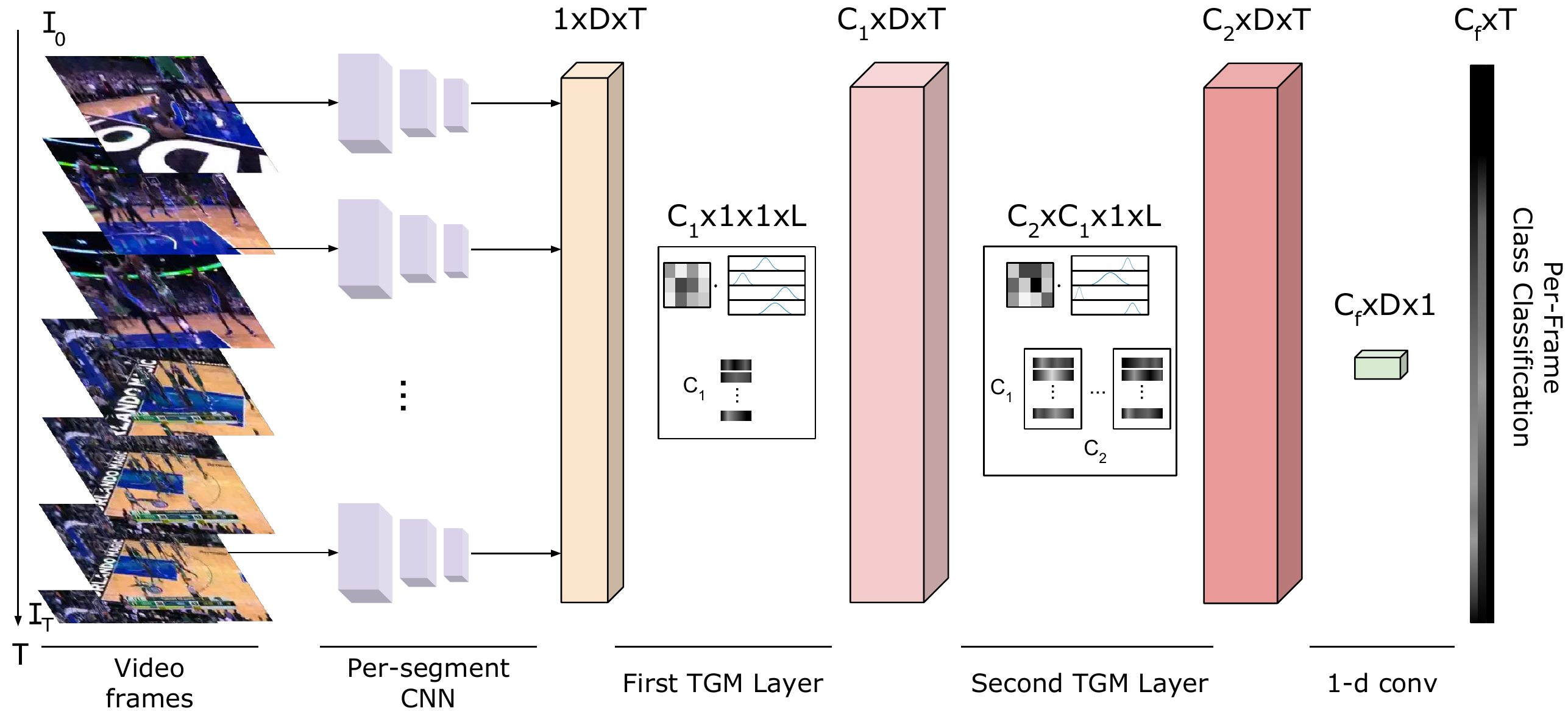}
    \caption{An overview of an example video CNN model with two TGM layers. Because of its fully convolutional design, it is able to handle videos with any length.}
    \label{fig:model-overview}
\end{figure*}

In the above subsection, we introduced an approach of stacking multiple TGM layers to model a hierarchical composition of temporal representations. However, in the grouped convolution case, each output channel of the layer is solely dependent on its corresponding input channel. That is, each kernel only considers information from a single output channel of the previous layer.

Therefore, we further generalize our TGM layer so that the layer combines representations from multiple input channels for each output channel while using the learned temporal kernels. We learn a set of convolutional kernels $\tens{K}\in\mathcal{R}^{C_{out}\times C_{in}\times L}$ (i.e., we learn $C_{out}\cdot C_{in}$ Gaussian mixtures). Given $\tens{f}$ which is the $C_{in}\times D\times T$ representation, for each output channel $i\in [1,C_{out}]$ and each input channel $j\in[1,C_{in}]$ pair, we convolve the associated filters with the input.
\begin{equation}
    \bm{G}_{i,j} = (\tens{f}_j * \bm{K}_{i,j})
\end{equation}
where each $\bm{G}_{i,j}$ is a $D\times T$-dim representation.

We then learn a 1x1 convolution followed by a ReLU activation function for each $i\in [1,C_{out}]$, which we call $w_i$, that maps from $C_{in}$ channels to 1 channel. The 1x1 convolution learns to combine the channels from the previous layer. By design, the TGM kernel is positive and sums to 1. Adding the unconstrained 1x1 convolution adds non-linearity (using the ReLU activation function) to our layer while only adding $C_{out}\cdot C_{in}$ parameters. The layer is computed as:
\begin{equation}
  s_i = \bm{G}_{i} * w_i = (\tens{f}_j * \bm{K}_{i,j}) * w_i, ~~ \tens{S}=[s_1,s_2\ldots,s_{C_{out}}]
\end{equation}

We then stack the $s_i$ representations along the channel axis to produce $\tens{S}$, the $C_{out}\times D\times T$-dim representation. This process is illustrated in Fig.~\ref{fig:shared}. This method generalizes our approach to allow the layer to take input of $C_{in}\times D\times T$ and produce output of $C_{out}\times D\times T$. These layers can easily be stacked to learn a hierarchical representation.

\subsection{Video CNN models with TGM layers}
\label{subsec:model}

Our goal is to do activity detection which we define as making a per-frame (or per-segment) classification. Given a video, at each time step $t$, we want to make the model decide which activity the frame corresponds to (including no-activity). As a baseline, we train a fully-connected layer that classifies each per-frame $D$-dimensional vector, $v_t$. As multiple activities can occur at the same time, or no activities at all, we treat this as a mutli-label classification task. We minimize binary cross entropy:
\begin{equation}
    L(v) = \sum_{t,c} z_{t,c}\log(p(c|v_t)) + (1-z_{t,c})\log(1-p(c|v_t))
\end{equation}
where $z_{t,c}$ is the ground truth label, 1 if activity $c$ is occurring at time $t$ and $p(c|v_t)$ is the output of our model for class $c$ at time $t$. Fig. \ref{fig:model-overview} shows an example CNN.

\section{Experiments}

\subsection{Implementation and baselines}
\label{subsec:implementation}

\paragraph{Implementation}

We used I3D~\citep{carreira2017quo} and the two-stream version of InceptionV3~\citep{szegedy2016rethinking} pretrained on Imagenet and Kinetics as our base per-frame CNNs. Our default $L$ setting used for the TGM layers as well as the other baselines was as follows: when using I3D segment features (3 features per second from 24fps videos), the 1 layer models used $L=15$ and the 3 layer models used $L=5$. When using InceptionV3 frame feature (at 8 fps), the 1 layer models used $L=30$ and the 3 layer models used $L=10$. These layers were attached on top of the base CNN, as described in Subsection \ref{subsec:model}. Please check Appendix for implementation and training details and results on other datasets.

\paragraph{Baselines}
\begin{table*}[t]
    \centering
    \caption{Comparison of various architectures on MultiTHUMOS using both I3D per-segment and InceptionV3 per-frame features. We found that TGM layers with 1x1 convolution channel combination performed the best. Results are in mAP. Note that we use the same filter length for ``Temporal Conv'' and ``TGM'' models, as described in Section \ref{subsec:implementation}.}
    \small
    \begin{tabular}{c|ccc||ccc}
        \toprule
                         &  \multicolumn{3}{c||}{I3D} & \multicolumn{3}{c}{InceptionV3}\\
                         & Spatial  & Temporal  & Two-Stream  & Spatial & Temporal & Two-Stream \\
        \midrule
        Baseline         & 22.3 & 25.0 & 29.7      & 13.6 & 14.1 & 15.2 \\
        Temporal Conv    & 32.5 & 35.5 & 38.4      & 15.2 & 15.5 & 15.8 \\
        3 Temporal Conv  & 20.4 & 23.4 & 24.4      &  5.3 &  6.1 & 6.5 \\
        \midrule
        \multicolumn{7}{l}{TGM layers with grouped convolution}\\
        \midrule
        1 TGM            & 35.1 & 37.8 & 40.5      & 16.3 & 17.5 & 18.0 \\
        3 TGM            & 36.4 & 42.3 & 43.5 & 17.5 & 18.3 & 19.2 \\  
        \midrule
        \multicolumn{7}{l}{TGM layers with channel combination}\\
        \midrule
        1 TGM (soft)   & 35.2 & 37.9 & 40.2     & 17.2 & 17.6 & 18.4 \\
        1 TGM (1x1)    & 36.1 & 38.2 & 40.8     & 17.2 & 17.7 & 18.4 \\
        3 TGM (soft)   & 36.2 & 40.1 & 42.3     & 17.5 & 19.1 & 21.2 \\
        3 TGM (1x1)    & 37.2 & 42.1 & 44.3     & 17.9 & 19.3 & 22.2 \\
        5 TGM (1x1)    & 37.4 & 42.4 & \textbf{44.8}     & 18.2 & 20.3 & 23.4 \\
        \bottomrule
    \end{tabular}
    \label{tab:multithumos-results}
\end{table*}

In order to confirm the advantages of our TGM layers, particularly against previous temporal models, we implemented many baselines. The first is \textbf{(i)} a standard per-frame classifier in which the prediction at each time-step only depends on a single feature vector with no contextual temporal information. We also used \textbf{(ii)} LSTMs on top of per-frame representations, which were popularly used to capture temporal information~\citep{donahue2015long}. We train a bi-directional LSTM with 512 hidden units to make per-frame predictions. We also tried \textbf{(iii)} the fixed pyramid temporal max-pooling of level 3 \citep{ryoo2015pooled}.

We also compare our model against \textbf{(iv)} the model with standard temporal convolutional layers (i.e., 1-D convolution with a $D\times L$ kernel) on top of per-frame representations. This is similar to the temporal conv. used in \citep{tran18closer}. Temporal lengths (i.e., $L$) of the 1-D conv. filters and the pooling windows were set to be identical to the TGM filters. That is, they capture the same temporal duration as TGMs. In all our experiments, we follow the standard evaluation setting of computing per-frame mean average precision (mAP) and report those values. 

We also compare to many different forms of the TGM layer: \textbf{(v)} 1-D convolution with a single Gaussian mixture per input (Fig. \ref{fig:tgm-shared-1d}); \textbf{(vi)} 1-D convolution with the kernel consisting of many Gaussian mixtures (Fig. \ref{fig:tgm-as-1d}); \textbf{(vii)} our TC-grouping with unconstrained kernels (see Appendix for figure); \textbf{(viii)} with a learned mixture of random temporal filters and \textbf{(ix)} with a learned mixture of fixed Gaussians.

In addition, we also tried the approach of combining our TGM layers with the super-event representations, which capture global context \citep{piergiovanni2018super}. We concatenated the learned super-event representation with our representations from TGM layers.

\subsection{THUMOS / MultiTHUMOS}

\paragraph{Dataset}

We conducted our experiments on both THUMOS~\citep{THUMOS14} and MultiTHUMOS~\citep{yeung2015every} datasets, while using the more challenging MultiTHUMOS as the main dataset. MultiTHUMOS is an extended version of the THUMOS dataset that densely annotates the continuous videos. The dataset consists of 65 different classes, compared to 20 in THUMOS, and contains on average 10.5 activities per video and 1.5 labels per frame and up to 25 activity instances in each video. This is in contrast to many other activity detection datasets such as ActivityNet~\citep{caba2015activitynet}, which only has on average $\sim$1 activity per video. THUMOS and MultiTHUMOS consists of YouTube videos of various sport activities like basketball/volleyball games, weight lifting, and track/field.

We followed the standard evaluation setting of each dataset (e.g., measuring per-frame mAP in MultiTHUMOS and using IoU=0.5 in THUMOS-14). There are 1010 validation videos and 1574 test videos. We used these continuous validation videos for the model training. We did not take advantage of the separate training set with segmented videos; even without them, we outperformed the state-of-the-arts.

\vspace{-4pt}
\paragraph{Results}
We compared baselines as well as multiple different versions of our architectures, shown in Table~\ref{tab:multithumos-results}. 
The model with our TGM layers consistently outperformed baseline I3D (or InceptionV3) while using the same per-segment representations.
Learning 3 TGM layers further improved the performance. We also note that the use of a 1x1 convoltuion + ReLU non-linearity improves performance more than soft-attention (weighted averaging), confirming that this additional non-linearity is beneficial. On the other hand, we found that stacking multiple standard temporal convolutional layers does not improve performance, often performing worse than the baseline. While a single standard temporal conv. layer improves over the baseline, stacking multiple layers significantly increases the number of parameters to learn (Table \ref{tab:parameters}) and we suspect this causes overfitting with the limited amount of samples in the dataset.

\begin{table}
\caption{Additional number of parameters for models when added to the base architecture (e.g., I3D or Inception V3).}
\label{tab:parameters}
\centering
\begin{tabular}{c|c}
\toprule
Model & \# of parameters \\
\midrule
LSTM              & 10.5M \\
1 Temporal Conv   & 10.5M \\
3 Temporal Conv   & 31.5M \\
1 TGM Layer       & 10K \\
3 TGM Layers      & 100K \\
5 TGM Layers      & 200K \\
\bottomrule
\end{tabular}
\end{table}

In Table \ref{tab:results-other}, we explicitly compare the results of using an LSTM or temporal conv. with a similar number of parameters to our TGM. This was done by making their temporal conv. filters to share values across multiple channels. These models result in nearly random performance, as they were not designed to cope with a small number of parameters.
We also show results with a mixture of random (fixed) temporal filters and with a mixture of fixed Gaussians. These results confirm that (i) modeling the temporal structure as a learned Gaussian mixture is beneficial and that (ii) further learning the Gaussian distribution parameters is important.

Also, in Table \ref{tab:layer-types}, we compare to the different forms of temporal convolution (as illustrated in Fig. \ref{fig:various-baselines}). We find that each filter using one Gaussian mixture across all channels is not beneficial, while using a Gaussian mixture per-input channel (i.e., standard 1-D conv. with Gaussian mixture kernels) outperforms standard 1-D conv. Further, we find that using TC-grouping outperforms standard 1-D conv even with unconstrained kernels, although this itself is not as effective as 1-D conv. with Gaussian mixtures. Finally, we find that our designed TGM layer performs the best, confirming that both modeling temporal information as Gaussian mixtures and our designed TC-grouping are useful for activity detection.




\begin{table}
\caption{Comparison of previous methods with comparable number of parameters and random forms of our TGM layer (using two-stream I3D on MultiTHUMOS).}
\label{tab:results-other}
\centering
\begin{tabular}{c|c}
\toprule
Model & mAP \\
\midrule
LSTM with 100k parameters    & 6.5\\
Temporal Conv. with 100k parameters & 7.3\\
TGM with random temporal filters & 34.5\\
TGM with fixed Gaussians  & 38.5\\
Full TGM   & \bf{44.3}\\
\bottomrule
\end{tabular}
\end{table}

\begin{table}
\caption{Comparison of the different forms of temporal convolution on MultiTHUMOS using RGB I3D features. We set $L=15$ and used 1 layer models for these experiments.}
\label{tab:layer-types}
\centering
\setlength\extrarowheight{0pt}
\begin{tabular}{c|c}
\toprule
Standard 1-D Convolution (Fig. \ref{fig:conv1d})      & 32.5  \\
1-D Conv with shared Gaussian mixture (Fig. \ref{fig:tgm-shared-1d})         & 28.6  \\
1-D Conv with Gaussian mixtures (Fig. \ref{fig:tgm-as-1d})      & 33.2 \\
TC-grouping with unconstrained kernel     & 32.8   \\
Our TGM Layer   & \textbf{36.1} \\
\bottomrule
\end{tabular}
\end{table}

Learning multiple TGM layers with channel combination outperforms the grouped convolution version of TGM and all the baselines. We also experimented with a version using soft-attention weights to combine the TGM layer channels, in addition to our method (Fig.~\ref{fig:shared}) of using 1x1 convolution followed by a ReLU (to gain non-linearity). We found that the 1x1 convolution performed better. We tested various number of Gaussian mixtures (i.e., output channels) and found that using 80 for the first and second layer and using 65 (i.e., number of classes) for the final layer performs best.



\begin{figure}
    \centering
    \includegraphics[width=\linewidth]{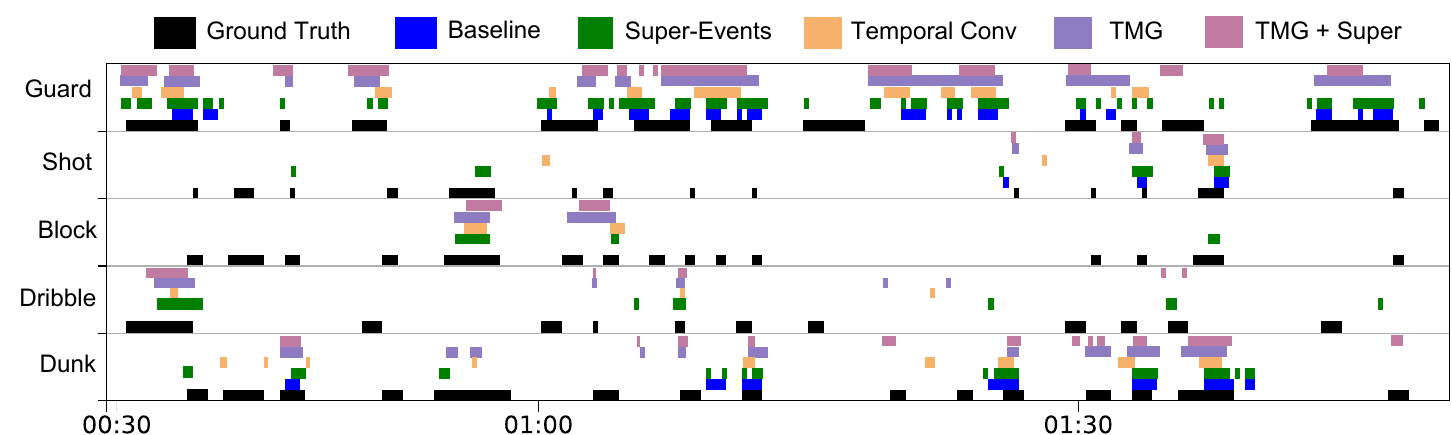}
    \caption{Illustration of the temporal regions classified as various basketball activities from a basketball game video in MultiTHUMOS. The TGM layers greatly improve performance.}
    \label{fig:multithumos-res}
\end{figure}

Tables~\ref{tab:multithumos} and \ref{tab:thumos} compare our model using TGM layers with multiple previous state-of-the-art approaches and baselines. Our approach meaningfully outperforms all previous approaches. Importantly, we are comparing our approach with different methods of capturing temporal information such as LSTMs and fixed temporal pyramid pooling while making them use the exactly same per-frame representations. We found that while all these methods capture some temporal information, the TGM layers provide the best performance. Further, combining the super-event representation~\citep{piergiovanni2018super} with our TGM feature also benefited detection, confirming that the TGMs and super-events capture different aspects of the activity videos. In Fig.~\ref{fig:multithumos-res}, we show an example of the various models predictions on a basketball video. We outperform the previous state-of-the-art performance (mAP) by 10\% (36.4 vs. 46.4).


\begin{table}
\caption{Performances of the state-of-the-art methods and our approach on MultiTHUMOS. Our approach meaningfully outperforms all previous results.}
\label{tab:multithumos}
\centering
\setlength\extrarowheight{0pt}
\begin{tabular}{c|c}
\toprule
 & mAP \\
\midrule
Two-stream~\citep{yeung2015every}    & 27.6\\
Two-stream + LSTM~\citep{yeung2015every}          & 28.1\\
Multi-LSTM~\citep{yeung2015every}    & 29.6\\
Predictive-corrective~\citep{dave2017predictive} & 29.7\\
SSN \cite{zhao2017temporal} & 30.3 \\
I3D baseline                         & 29.7 \\
I3D + LSTM                           & 29.9 \\
I3D + temporal pyramid               & 31.2 \\
I3D + super-events~\citep{piergiovanni2018super}            & 36.4 \\
I3D + our TGMs                       & 44.3 \\
I3D + super-events + our TGMs        & \bf{46.4} \\
\bottomrule
\end{tabular}
\end{table}

\begin{table}
\caption{Performances of our approach compared to the state-of-the-arts on the continuous version of THUMOS-14, with IoU=0.5.}
\label{tab:thumos}
\centering
\begin{tabular}{c|c}
\toprule
 & mAP \\
\midrule
R-C3D \cite{xu2017r}                                 &  28.9 \\
SSN \cite{zhao2017temporal} & 29.1 \\
TAL-Net \cite{chao2018rethinking} & 42.8 \\
I3D Baseline & 43.5 \\
I3D + super-events\citep{piergiovanni2018super}                   & 47.8 \\
I3D + 1 TGMs                                   & 45.2  \\
I3D + 3 TGMs                                   & 53.5 \\
I3D + 3 TGMs + super-events                     &  57.0 \\
\bottomrule
\end{tabular}
\end{table}

\subsection{Charades}

\vspace{-3pt}
\paragraph{Dataset} Charades~\citep{sigurdsson2016hollywood} is a large scale dataset with 9848 videos across 157 activity classes. These videos were recorded in home environments of the participants based on provided scripts. Each video contains on an average of 6.8 activity instances, and there are often complex activities co-occurring. The activities were mainly performed at home. For example, some activity classes are `preparing a meal', `eating', `sitting', `cleaning', etc. 

In our experiments, we follow the original Charades detection setting (i.e., Charades\_v1\_localize evaluation), which is the original setting used in many previous approaches \citep{sigurdsson2016asynchronous,piergiovanni2018super}. 

\vspace{-4pt}
\paragraph{Results}
We compare our results with the state-of-the-arts in Table~\ref{tab:charades-comp}. To our knowledge, our method is obtaining the best known performance in the original localization setting of the Charades dataset. Notably, it is performing better than I3D that obtained the best competition performance, while using the same feature. Our method also outperforms standard temporal convolution, LSTMs, and fixed pyramid pooling, as well as the use of latent super-events.
When setting $L=30$ and using 3 TGM layers, our model is able to capture around 800 frames (about $\pm$15 seconds from each frame) of temporal information, significantly more than previous works (e.g., I3D only captures $\pm$2 seconds). This confirms that a key advantage of the TGM layer is that the number of parameters is independent of the filter length.

\begin{table}
\caption{Per-frame mAP on Charades, evaluated with the `Charades\_v1\_localize' setting. I3D models are two-stream, using both RGB and optical flow inputs.} 
\label{tab:charades-comp}
\centering
\setlength\extrarowheight{0pt}
\begin{tabular}{c|c}
\toprule
 & mAP \\
\midrule
Predictive-corrective~\citep{dave2017predictive} & 8.9 \\
Two-stream~\citep{sigurdsson2016asynchronous}& 8.94 \\
Two-stream+LSTM~\citep{sigurdsson2016asynchronous}& 9.6 \\
R-C3D~\citep{xu2017r}                         &  12.7 \\
Sigurdsson et al.~\citep{sigurdsson2016asynchronous} & 12.8 \\
SSN \cite{zhao2017temporal} & 16.4 \\
I3D baseline                                 & 17.2 \\
I3D + 3 temporal conv. layers ($L=5$)                 &  17.5   \\
I3D + 3 temporal conv. layers ($L=30$)                 &  12.5   \\
I3D + LSTM                                   & 18.1 \\
I3D + fixed temporal pyramid                & 18.2 \\
I3D + super-events~\citep{piergiovanni2018super}                   & 19.4 \\
I3D + 3 TGMs ($L=5$)                         & 20.6  \\
I3D + 3 TGMs ($L=30$)                          & 21.5  \\
I3D + 3 TGMs ($L=5$)  + super-events                     & 21.8  \\
I3D + 3 TGMs ($L=30$) + super-events                     & \bf{22.3}  \\
\bottomrule
\end{tabular}
\end{table}

\begin{figure}
    \centering
    \includegraphics[width=\linewidth]{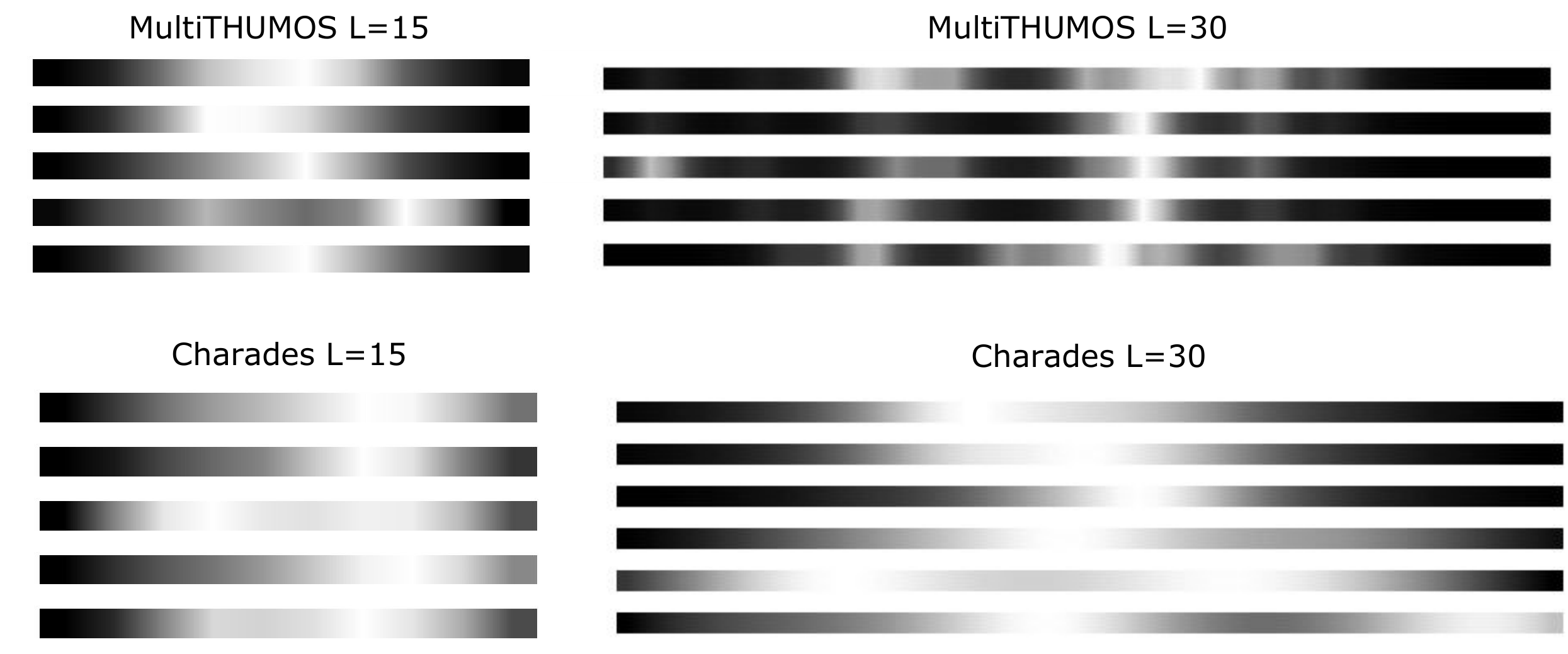}
    \caption{Illustration of several learned TGM kernels. On MultiTHUMOS, it learns to focus on shorter intervals to capture shorter events. On Charades, the Gaussians have a larger $\sigma$ value, resulting in filters that attend to longer temporal durations.}
    \label{fig:tgm-vis}
\end{figure}

\begin{table}
\caption{Effect of $L$ on MultiTHUMOS using only RGB I3D features. Note that the 3 TGM layer models capture information in larger temporal intervals than the 1 TGM layer models for the same values of $L$. We also compare to using standard one-layer 1-D conv layer with different values of $L$.}
\label{tab:l-exp-mt}
\centering
\begin{tabular}{c|ccc}
\toprule
 &  1 TGM & 3 TGM & 1-D Conv  \\
\midrule
I3D Baseline   & 22.3 & -      & -      \\
$L=3$      & 30.2 &  31.7      & 26.6   \\
$L=5$      & 32.5 &  \bf{37.2} & 28.3   \\
$L=10$     & 34.5 &  35.4      & 31.7   \\
$L=15$     & \bf{36.1} &  34.1 & 32.5   \\
$L=30$     & 32.5 &  33.9      & 26.5   \\
$L=50$     & 32.1 &  33.7      & 15.4  \\
\bottomrule
\end{tabular}
\end{table}

In Table \ref{tab:l-exp-mt} (and Table 3 in Appendix), we compare different values of $L$. Due to the short duration of activities in MultiTHUMOS (3.3 seconds), we find that $L=5$ with 3 layers performs the best.  Larger values of $L$ capture too much unneeded temporal information, but due to the Gaussian structure, it does not drastically harm performance. Figure \ref{fig:tgm-vis} shows that even with longer kernels, the Gaussians learn to focus mostly on the center of the interval and capture short intervals. Thus, having too long intervals does not drastically harm performance, which is in contrast to the standard 1-D convolution. Note that for Charades, the temporal kernels learned to capture much longer temporal durations, as the average activity in charades is 12.8 seconds and larger values of $L$ are better.

\vspace{-3pt}
\section{Conclusions}
\vspace{-4pt}

We newly introduced the Temporal Gaussian Mixture (TGM) layer and demonstrated its effectiveness for multi-activity detection in continuous videos. Our layer is fully differentiable and trainable using standard backpropagation, designed to learn temporal structure. We were able to confirm that our layer performs superior to state-of-the-art methods on activity detection datasets including MultiTHUMOS and Charades, obtaining the best known performance.  We also tested our approach with two more public video datasets, MLB-YouTube \citep{mlbyoutube2018} and AVA \citep{ava2017}, and confirmed its advantage over the previous works in Appendix.

\section*{Acknowledgements} This work was supported in part by the National Science Foundation (IIS-1812943 and CNS-1814985), and by the ICT R\&D program of MSIP/IITP, Republic of Korea (17ZF1200, Development of XDMedia Solution for Invigoration of Realistic Media Industry).

\bibliographystyle{icml2019}
\bibliography{egbib}

\clearpage
\newpage

\appendix

\section{Implementation Details}

As our base per-segment CNN, we use the I3D~\citep{carreira2017quo} network pretrained on the ImageNet and Kinetics~\citep{kay2017kinetics} datasets. I3D obtained state-of-the-art results on segmented video tasks, and this allows us to obtain reliable $v_t$. We also use two-stream version of InceptionV3~\citep{szegedy2016rethinking} pretrained on Imagenet and Kinetics as our base per-frame CNN, and compared them. We chose InceptionV3 as it is deeper than previous two-stream CNNs such as \citep{simonyan2014two,feichtenhofer2016convolutional}. We extracted frames from the videos at 25 fps, computed TVL1~\citep{zach2007duality} optical flow, clipped to $[-20,20]$. For InceptionV3, we computed features for every 3 frames (8 fps). For I3D, every frame was used as the input. I3D has a temporal stride of 8, resulting in 3 features per second (3 fps). By design, I3D has a temporal resolution of 99 frames, so each feature is able to capture up to 99 frames of temporal information.

We implemented our TGM layers as well as other baseline layers in PyTorch. Our default setting was as follows: for 3-layer models, we set $L=10$ for frame-based features (i.e., InceptionV3) and $L=5$ for segment-based features (i.e., I3D), as each segment already contains some temporal information. For 1-layer models, we set $L=30$ for frame-based features and $L=15$ for segment-based features. We set $M=16$ and $C_{out}=80$ and $C_{out}=65$ for the last TGM layer. We found these values to work well on a held out portion of the training set of MultiTHUMOS. In all models, we used one fully-connected layer at the end to make the per-frame or per-segment classification.

We trained our models using the Adam~\citep{kingma2014adam} optimizer with the learning rate set to 0.01. We decayed the learning rate by a factor of 10 after every 10 training epochs. We trained our models for 50 epochs. We plan to make all our source code and trained models publicly available once the paper is published.

\section{Hyperparameter Experiments}
\begin{table}
\begin{minipage}{0.48\textwidth}
\caption{Comparison of various values of $M$ on MultiTHUMOS and Charades using RGB I3D features. For these experiments, 1 layer was used with $L=15$ and $C_{out}=16$.}
\label{tab:M-exp}
\centering
\begin{tabular}{c|cc}
\toprule
 & MultiTHUMOS & Charades \\
\midrule
$M=2$      & 27.8 & 15.5  \\
$M=4$      & 33.1 & 16.2 \\
$M=8$      & 34.8 & 17.5  \\
$M=16$     & 36.1 & 17.5  \\
$M=32$     & 35.7 & 17.1  \\
$M=64$     & 35.8 & 17.3  \\
\bottomrule
\end{tabular}
\end{minipage}
\hfill
\begin{minipage}{0.48\textwidth}
\caption{Comparison of values of $C_{out}$ on MultiTHUMOS and Charades using RGB I3D features. For these experiments, 1 layer was used with $L=15$ and $M=16$.}
\label{tab:cout-exp}
\centering
\begin{tabular}{c|cc}
\toprule
 & MultiTHUMOS & Charades \\
\midrule
$C_{out}=1$      & 33.5 & 16.2  \\
$C_{out}=4$      & 34.2 & 17.4  \\
$C_{out}=8$      & 35.5 & 17.5  \\
$C_{out}=16$     & 36.1 & 17.5  \\
$C_{out}=32$     & 36.0 & 17.2  \\
$C_{out}=64$     & 36.1 & 17.4  \\
$C_{out}=80$     & 36.1 & 17.5  \\
\bottomrule
\end{tabular}
\end{minipage}
\end{table}
We conducted a set of experiments to compare the effects of the temporal duration, $L$, number of Gaussians, $M$, and the number of output channels, $C_{out}$. For these experiments, we only used the one-stream version of I3D with RGB inputs.

\begin{table*}
\caption{Effect of $L$ on MultiTHUMOS and Charades using only RGB I3D features. Note that the 3 TGM layer models have larger temporal resolution than the 1 TGM layer models for the same values of $L$. We also compare to using standard one-layer 1-D conv layer with different values of $L$.}
\label{atab:l-exp}
\centering
\begin{tabular}{c|ccc||ccc}
\toprule
 & \multicolumn{3}{c||}{MultiTHUMOS} & \multicolumn{3}{c}{Charades} \\
 &  1 Layer & 3 Layers & 1-D Conv & 1 Layer & 3 Layers & 1-D Conv \\
\midrule
I3D Baseline   & 22.3 & -      & -     & 15.3 & - & - \\
$L=3$      & 30.2 &  31.7      & 26.6  & 15.5 & 16.1 & 15.5 \\
$L=5$      & 32.5 &  \bf{37.2} & 28.3  & 15.7 & 17.8 & 16.3 \\
$L=10$     & 34.5 &  35.4      & 31.7  & 16.1 & 18.2 & 16.6 \\
$L=15$     & \bf{36.1} &  34.1 & 32.5  & 17.5 & 18.6 & 16.8 \\
$L=30$     & 32.5 &  33.9      & 26.5  & 18.1 & \bf{18.9} & 12.1 \\
$L=50$     & 32.1 &  33.7      & 15.4  & \bf{18.3} & 18.8 & 6.7 \\
\bottomrule
\end{tabular}
\end{table*}

\paragraph{Effect of $L$:} In Table \ref{atab:l-exp}, we compare different values of $L$. For these experiments, we use $M=16$ and $C_{out}=16$. We find that the 3-layer model with $L=5$ performs the best. With I3D features, this allows the model to capture up to 8 seconds of information. The average activity in MultiTHUMOS is 3.3 seconds long and the maximum is 14.7 seconds long, and with this setting, the model is able to capture enough temporal context to perform well. Larger values of $L$ capture too much temporal information, but due to the Gaussian structure, it does not drastically harm performance. Figure \ref{afig:tgm-vis} shows that even with longer kernels, the Gaussians learn to focus mostly on the center of the interval and capture the rough duration of the activities. Thus, having too long intervals does not drastically harm performance, which is in contrast to the standard 1-D convolution. Note that for Charades, the temporal kernels are learned to capture much longer temporal duration, as the average activity in charades is 12.8 seconds and larger values of $L$ perform better.

Figure \ref{afig:tgm-vis} illustrates examples of the learned TGM kernels of various lengths. The figure shows that the kernels focus on short temporal intervals on MultiTHUMOS even if we make the filters longer, as the activities are an average of 3.3 seconds long. On Charades, the TGM kernels learn to capture much longer intervals, as the activities are an average of 12.8 seconds long. We believe that this suggests TGMs are learning to capture information from the important necessary intervals.

In Table \ref{atab:l-exp}, we also report the results of using a standard 1-D conv. layer with different $L$ values. The number of parameters in our TGM layer is independent of $L$, however, with the standard 1-D conv. layer, the number of parameters increases as $L$ increases. We find that increasing $L$ with 1-D convolution helps for small values of $L$, but for $L > 15$, the performance drastically drops, while TGM layers only show a small decrease.

\paragraph{Effect of $M$:} In Table \ref{tab:M-exp}, we compare different values of $M$. For these experiments, we set $L=15$ and $C_{out}=16$. We find that $M=16$ performs best, suggesting that smaller values of $M$ restrict the possible temporal kernels too much. We also observe that larger values of $M$ performs slightly worse than $M=16$ (but not much), likely because they introduce more parameters than needed. When $M$ and $L$ have similar values, it allows the model to learn a sufficient number of Gaussians and create a diverse range of temporal kernels. When $M$ is larger than $L$, it results in learning a kernel similar to standard 1-D convolution.

\paragraph{Effect of $C_{out}$:} In Table \ref{tab:cout-exp}, we compare different values of $C_{out}$. For these experiments, $L=15$, we used 1-layer and $M=16$. We find that $C_{out}$ performs best when set to 16 or larger on these datasets. Larger values of $C_{out}$ seem to capture redundant information, as it does not lower performance.

\begin{figure*}
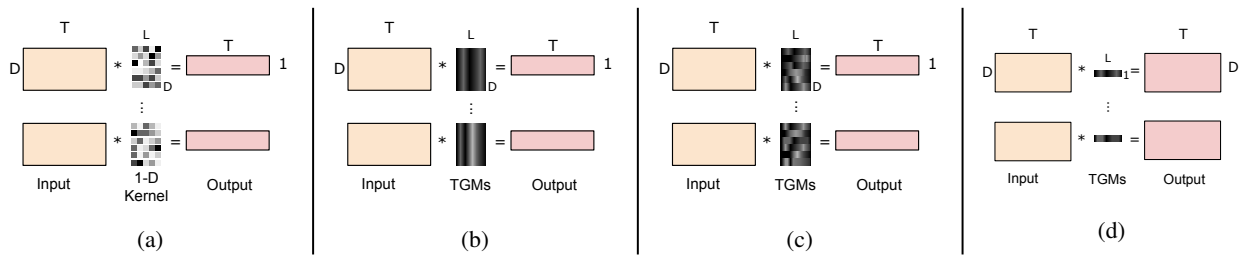

\centering
  \begin{subfigure}{.24\textwidth}
  \centering
    \includegraphics[width=\textwidth]{figures/1d-conv.pdf}
    \caption{}
    \label{afig:conv1d}
  \end{subfigure}\rulesep%
  \begin{subfigure}{.24\textwidth}
  \centering
    \includegraphics[width=\textwidth]{figures/tgm-shared-1d.pdf}
    \caption{}
    \label{afig:tgm-shared-1d}
  \end{subfigure}\rulesep%
  \begin{subfigure}{.24\textwidth}
  \centering
    \includegraphics[width=\textwidth]{figures/tgm-as-1d.pdf}
    \caption{}
    \label{afig:tgm-as-1d}
  \end{subfigure}\rulesep%
  \begin{subfigure}{.22\textwidth}
  \centering
    \includegraphics[width=\textwidth]{figures/tgm-draw2.pdf}
    \caption{}
    \label{afig:real-tgm}
  \end{subfigure}
  \caption{{\bf (a-c)} Different forms of 1-D temporal convolutions which take a $D\times T$ input and produces a $C\times T$ output based on $C$ number of $D\times L$ kernels: {\bf (a)} the standard 1-D convolution, {\bf (b)} using Gaussian mixtures for 1-D convolution while sharing Gaussian mixtures across input channels, and {\bf (c)} using $D$ different Gaussian mixtures for 1-D convolution. {\bf (d)} Our TGM layer in its simplest form (i.e., 1-layer case) applying the $1\times L$ temporal kernel in a 2-D convolutional fashion, maintaining both time and feature axis.}
  \label{afig:various-baselines}
\end{figure*}

\begin{figure*}
    \centering
    \includegraphics[width=\textwidth]{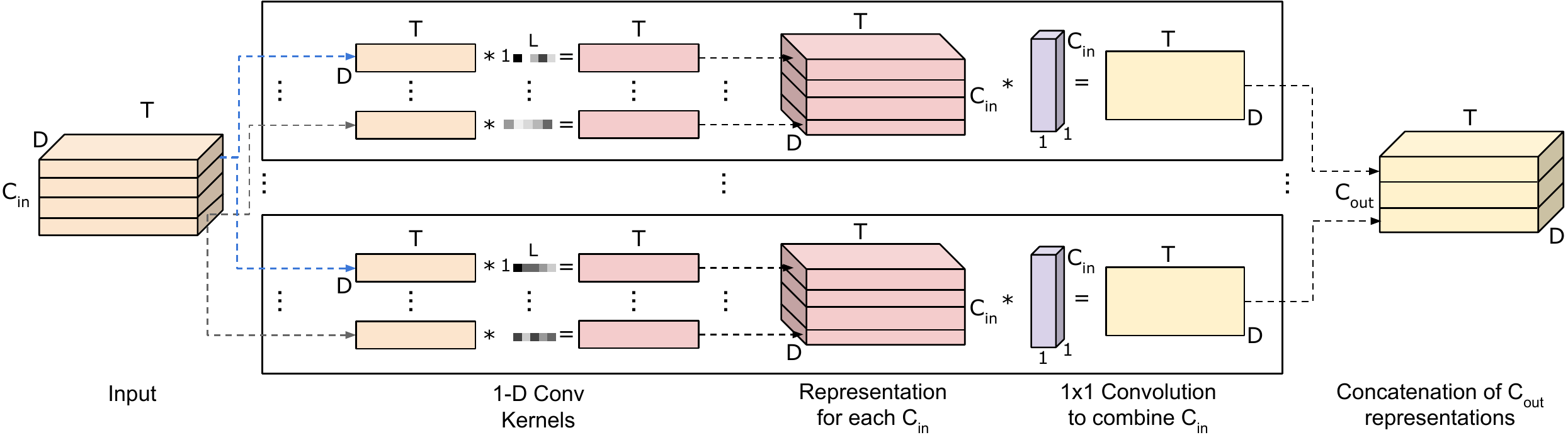}
    \caption{A temporal convolutional layer with channel combination similar to Fig. 4 (in main paper). The difference is that this layer does not learn Gaussian mixtures, but unconstrained 1-D temporal kernels.}
    \label{fig:tgm-with-1d}
\end{figure*}


\section{Comparison of Different Layer Forms}
\begin{table}
\caption{Comparison of the different forms of temporal convolution on MultiTHUMOS using RGB I3D features. We set $L=15$ and used 1 layer models for these experiments.}
\label{atab:layer-types}
\centering
\begin{tabular}{c|c}
\toprule
Standard 1-D Convolution (Fig. \ref{afig:conv1d})      & 32.5  \\
1-D Conv with 1 Gaussian (Fig. \ref{afig:tgm-shared-1d})         & 28.6  \\
1-D Conv with many Gaussians (Fig. \ref{afig:tgm-as-1d})      & 33.2 \\
TC-Conv with unconstrained kernel (Fig. \ref{fig:tgm-with-1d})     & 32.8   \\
Our TGM Layer   & \textbf{36.1} \\
\bottomrule
\end{tabular}
\end{table}

To confirm the various aspects of our design, we conducted experiments comparing different types of temporal convolution. In Fig. \ref{afig:conv1d} we illustrate the standard 1-D convolution, taking $D\times T$ input and producing a $C\times T$ output, where $D$ is the number of input channels and $C$ is the number of output channels. In Fig. \ref{afig:tgm-shared-1d}, we illustrate the method of applying a Gaussian mixture kernel as 1-D convolution. Here, the Gaussian mixture kernel is shared by all $D$ input channels and we learn a $C$ number of such kernels. In Fig. \ref{afig:tgm-as-1d}, we illustrate the approach of applying a Gaussian mixture kernel as 1-D convolution while learning $D$ different Gaussian mixtures. This is very similar to the standard 1-D convolution, except that the filter values are constrained to have the shape of Gaussian mixtures.

Fig. \ref{fig:tgm-with-1d} illustrates one more baseline. This is similar to our full TGM layer with the channel-combination (Fig. 4 in main paper). However, in this baseline, instead of learning Gaussian mixtures, we learn $C_{in}\cdot C_{out}$ number of $1\times L$ kernels. The kernel values are left unconstrained.
While the TGM layer has $2\cdot M + C_{in}\cdot C_{out}\cdot M + C_{in}\cdot C_{out}$ parameters, this layer has $L \cdot C_{in}\cdot C_{out}\cdot M + C_{in}\cdot C_{out}$, which is more than the TGM layer.

In Table \ref{atab:layer-types}, we compare the results of the various above-mentioned layers on MultiTHUMOS using RGB I3D features. We find that the Fig. \ref{afig:tgm-shared-1d} method performs poorly, while the Fig. \ref{afig:tgm-as-1d} method slightly outperforms the standard 1-D convolution.
The Fig. \ref{fig:tgm-with-1d} method is slightly better than the standard 1-D convolution, but performs worse than Fig. \ref{afig:tgm-as-1d}.
However, none of these layers perform as well as our TGM layer, confirming that both the design of learning Gaussian mixtures and maintaining temporal channel axis are important for activity detection.

\begin{figure*}
    \centering
    \includegraphics[width=\textwidth]{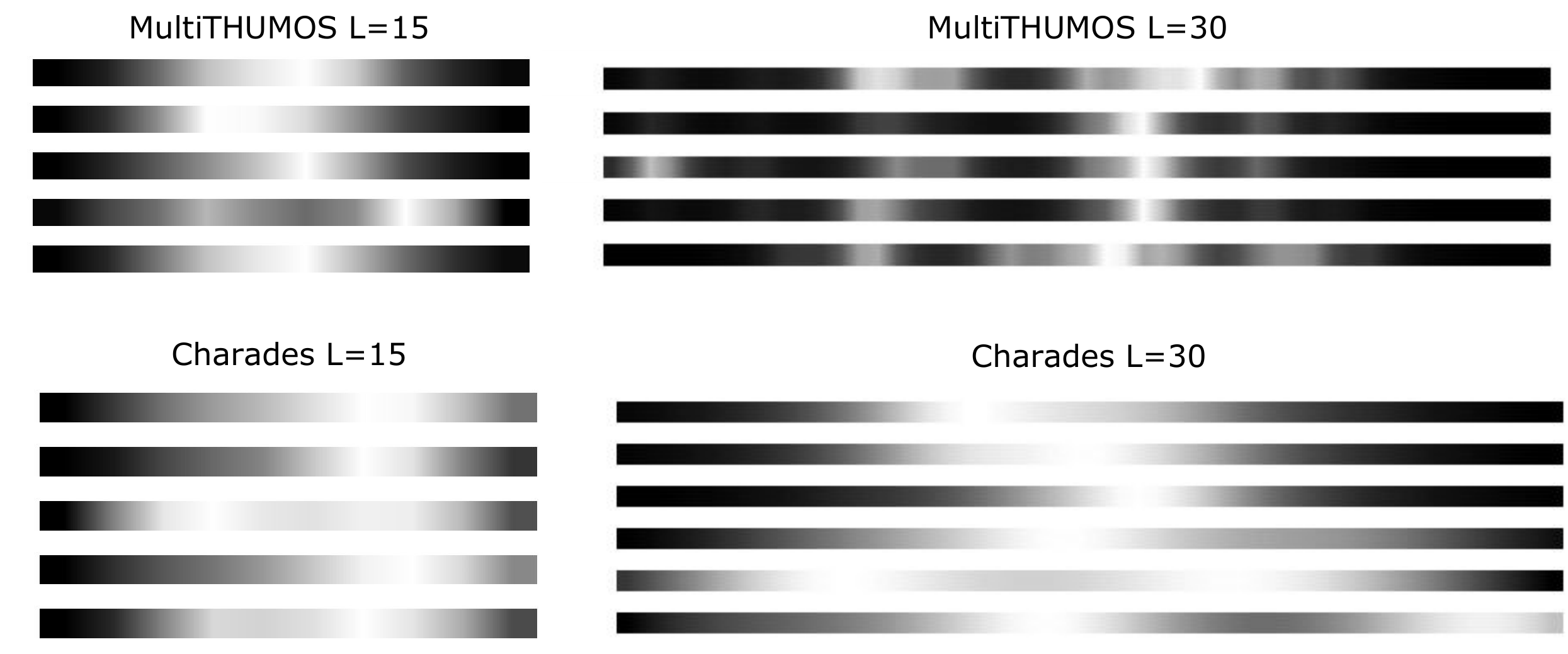}
    \caption{Illustration of several learned TGM kernels. On MultiTHUMOS, it learns to focus on shorter intervals to capture shorter events. On Charades, the Gaussians have a larger $\sigma$ value, resulting in filters that attend to longer temporal durations.}
    \label{afig:tgm-vis}
\end{figure*}

\section{Experiments on Additional Datasets}
\subsection{MLB-YouTube Dataset}
\begin{figure*}
    \centering
    \includegraphics[width=\textwidth]{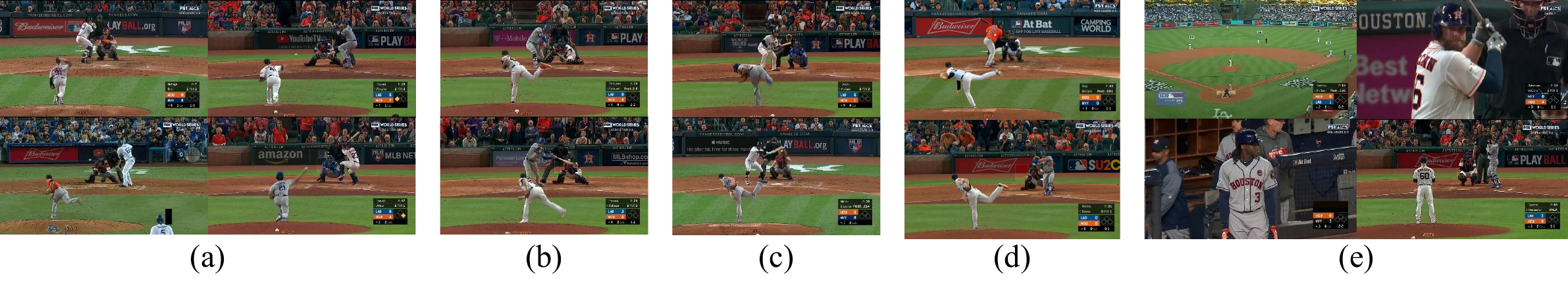}
    \caption{Examples of several of the activities in the MLB-YouTube dataset: (a) Pitch, (b) Hit, (c) Bunt,  (d) Hit by pitch, (e) No activity. This shows the difficulty of this dataset, as the difference between hit and bunt, swing and no swing are very small.}
    \label{fig:baseball-example}
\end{figure*}

\subsubsection{Dataset}
The MLB-YouTube dataset \citep{mlbyoutube2018} consists of 20 baseball games from the 2017 MLB post-season available on YouTube. This dataset consists of over 42 hours of video. For these experiments, we used the continuous video setting which have 2,126 1-2 minute long clips. Each clip is densely annotated with the baseball activities that occur. There are 8 activity classes: pitch, strike, ball, swing, hit, foul, hit by pitch, and bunt. Examples of some of these classes are shown in Fig.~\ref{fig:baseball-example}. Each continuous clip contains on average of 7.2 activities, giving a total of over 15,000 activity instances in the dataset.

What makes this dataset challenging is that the variation between classes is very small. In ActivityNet~\citep{caba2015activitynet}, for example, the difference between swimming and brushing hair is drastic. The background, motion, and even size of the person in the video is different. However, in broadcast baseball videos, the difference between a ball and a strike, or a swing and a bunt, are small. All actions are recorded from the same camera angle as we can confirm from Fig.~\ref{fig:baseball-example}.

\subsubsection{Results}
In Table~\ref{tab:mlb-youtube}, we compare various approaches on this dataset. Our TGM layers improve over the baseline by $\sim$6\% (40.1 vs. 34.2). Additionally, we compare to methods using the super-event representation~\citep{piergiovanni2018super}, which previously achieved state-of-the-art performance on several activity detection datasets. On this dataset, our approach outperforms the super-event representation, and further the concatenation of our TGM representation with such super-event representation performs best by a significant margin ($\sim$13\% compared to the baseline). This suggests that TGMs and super-event capture different temporal information and are both useful to the detection task.

We further find that using multiple, standard temporal convolution layers leads to worse performance, likely due to overfitting from the large number of parameters. While using multiple TGM layers improves performance, confirming that the Gaussian structure and sparsity constraint benefits model learning.

\begin{table*}
\caption{Result mAP on the MLB-YouTube dataset using InceptionV3 and I3D to obtain features. Our TGM layers significantly outperform the baseline models. }
\label{tab:mlb-youtube}
\centering
\begin{tabular}{c|ccc}
\toprule
Model                            & Spatial   & Temporal  & Two-stream \\
\midrule
Random                           & 13.4 & 13.4 & 13.4  \\
\midrule
InceptionV3                      & 31.2 & 31.8 & 31.9 \\
InceptionV3 + LSTM               & 32.1 & 33.5 & 34.1 \\
InceptionV3 + 1 temporal conv               & 32.8 & 34.4 & 35.2 \\
InceptionV3 + 3 temporal conv               & 28.4 & 29.8 & 30.1 \\
InceptionV3 + super-events       & 31.5 & 36.2 & 39.6 \\
InceptionV3 + 1 TGM              & 32.4 & 36.3 & 37.4 \\
InceptionV3 + 3 TGM              & 33.2 & 38.2 & 38.2 \\
InceptionV3 + 3 TGM+super-events & 34.6 & 42.4 & 42.9 \\
\midrule
I3D                      & 33.8 & 35.1 & 34.2 \\
I3D + LSTM               & 36.2 & 37.3 & 39.4 \\
I3D + 1 temporal conv               & 37.3 & 38.6 & 39.9 \\
I3D + 3 temporal conv               & 32.4 & 34.6 & 35.6 \\
I3D + super-events       & 38.7 & 38.6 & 39.1 \\
I3D + 1 TGM              & 35.5 & 37.5 & 38.5 \\
I3D + 3 TGM              & 36.5 & 38.4 & 40.1  \\
I3D + 3 TGM+super-events & 39.4 & 46.0 & \bf{47.1} \\
\bottomrule
\end{tabular}
\end{table*}

\begin{table}
\caption{Results on AVA dataset with the temporal annotation-only setting (i.e., frame classification without using bounding box training labels).}
\label{tab:ava}
\centering
\begin{tabular}{c|c}
\toprule
 & mAP \\
\midrule
Random & 2.65 \\
I3D baseline                                 & 7.5 \\
I3D + 3 temporal conv. layers                 & 7.9    \\
I3D + LSTM                                   & 7.8 \\
I3D + super-events\citep{piergiovanni2018super}                   & 9.8 \\
I3D + 1 TGMs                                   & 11.2  \\
I3D + 3 TGMs                                   & 14.5 \\
I3D + 3 TGMs + super-events                     & 14.9  \\
\bottomrule
\end{tabular}
\end{table}

\subsection{AVA}
\subsubsection{Dataset}
AVA \citep{ava2017} is a large-scale video dataset containing of 80 atomic action classes in 57k video clips. These clips are drawn from movies. Existing datasets, such as Charades, have very specific actions that depend on objects, such as holding a cup vs. holding a picture. In AVA, the actions are intentionally generic, such as sit, stand, hold, carry, etc. Further, the AVA dataset is annotated with both spatial and temporal locations of activities. Since we are interested in temporal activity detection, we follow the setting of \citet{piergiovanni2018super} and label each frame with the occurring activities while ignoring the spatial location. We evaluate performance following the same method as MultiTHUMOS, Charades and MLB-YouTube by measuring per-frame mAP.

\subsubsection{Results}
In Table \ref{tab:ava}, we present the results of our model. We again find that temporal convolution and LSTMs provide some benefit over the baseline, but TGM layers further improve performance. Again, combining the TGM, which captures local temporal structure, with super-events which capture global temporal structure, provides the best performance by $\sim7.4\%$.




\subsection{Context Gaiting}

Context gating \cite{miech2017learnable} is an layer designed to capture relationships between network activations. However, it is designed for segmented video clip classification, as it originally takes a fixed-size input. Applying it to variable length continuous videos in a sliding-window fashion is possible, and we conducted this experiment with a window size of 30 (same temporal resolution as ours). When context gating is applied on top of I3D features, it gives 35.8 on MultiTHUMOS, lower than ours (44.3).

\end{document}